\PassOptionsToPackage{unicode}{hyperref}
\PassOptionsToPackage{hyphens}{url}
\PassOptionsToPackage{dvipsnames,svgnames*,x11names*}{xcolor}
\documentclass[11pt]{article}
\usepackage{lmodern}
\usepackage{setspace}
\usepackage{amssymb,amsmath}
\usepackage{subfig}
\usepackage{ifxetex,ifluatex}
\ifnum 0\ifxetex 1\fi\ifluatex 1\fi=0 
  \usepackage[T1]{fontenc}
  \usepackage[utf8]{inputenc}
  \usepackage{textcomp} 
\else 
  \usepackage{unicode-math}
  \defaultfontfeatures{Scale=MatchLowercase}
  \defaultfontfeatures[\rmfamily]{Ligatures=TeX,Scale=1}
\fi
\IfFileExists{upquote.sty}{\usepackage{upquote}}{}
\IfFileExists{microtype.sty}{
  \usepackage[]{microtype}
  \UseMicrotypeSet[protrusion]{basicmath} 
}{}
\makeatletter
\@ifundefined{KOMAClassName}{
  \IfFileExists{parskip.sty}{%
    \usepackage{parskip}
  }{
    \setlength{\parindent}{0pt}
    \setlength{\parskip}{6pt plus 2pt minus 1pt}}
}{
  \KOMAoptions{parskip=half}}
\makeatother
\usepackage{xcolor}
\usepackage{soulutf8}
 \usepackage{natbib}
\IfFileExists{xurl.sty}{\usepackage{xurl}}{} 
\IfFileExists{bookmark.sty}{\usepackage{bookmark}}{\usepackage{hyperref}}
\hypersetup{
  pdftitle={Look Who’s Talking: Interpretable Machine Learning for
Assessing Italian SMEs Credit Default},
  pdfauthor={Lisa Crosato, Caterina Liberati and Marco Repetto ~},
  colorlinks=true,
  linkcolor=blue,
  filecolor=Maroon,
  citecolor=blue,
  urlcolor=blue,
  pdfcreator={LaTeX via pandoc}}
\usepackage[margin=1in]{geometry}
\usepackage{longtable,booktabs}
\usepackage{etoolbox}
\makeatletter
\patchcmd\longtable{\par}{\if@noskipsec\mbox{}\fi\par}{}{}
\makeatother
\IfFileExists{footnotehyper.sty}{\usepackage{footnotehyper}}{\usepackage{footnote}}
\makesavenoteenv{longtable}
\usepackage{graphicx}
\makeatletter
\def\maxwidth{\ifdim\Gin@nat@width>\linewidth\linewidth\else\Gin@nat@width\fi}
\def\maxheight{\ifdim\Gin@nat@height>\textheight\textheight\else\Gin@nat@height\fi}
\makeatother
\setkeys{Gin}{width=\maxwidth,height=\maxheight,keepaspectratio}
\makeatletter
\def\fps@figure{htbp}
\makeatother
\providecommand{\tightlist}{%
  \setlength{\itemsep}{0pt}\setlength{\parskip}{0pt}}
\usepackage{pdflscape}
\usepackage{booktabs}
\usepackage{longtable}
\usepackage{array}
\usepackage{multirow}
\usepackage{wrapfig}
\usepackage{float}
\usepackage{colortbl}
\usepackage{pdflscape}
\usepackage{tabu}
\usepackage{threeparttable}
\usepackage{threeparttablex}
\usepackage[normalem]{ulem}
\usepackage{makecell}
\usepackage{xcolor}
\newlength{\cslhangindent}
  {\setlength{\parindent}{0pt}%
  \everypar{\setlength{\hangindent}{\cslhangindent}}\ignorespaces}%
  {\par}

\title{Look Who’s Talking: Interpretable Machine Learning for Assessing Italian SMEs Credit Default}
\author{Lisa Crosato\footnote{Department of Economics and 
		Bliss - Digital Impact Lab, 
		Ca' Foscari University of Venice, Cannaregio 873, 30121, Venice, 
		Italy,
		\nolinkurl{lisa.crosato@unive.it}}, Caterina 
	Liberati\footnote{Department of 
		Economics, Management and 
		Statistics (DEMS) and Center for European Studies (CefES), Bicocca 
		University, 
		Milano, Piazza 
		dell'Ateneo Nuovo, 1, 20126 Milan, Italy, 
		\nolinkurl{caterina.liberati@unimib.it}} and Marco 
	Repetto\footnote{Department of 
		Economics, Management and 
		Statistics (DEMS) Bicocca 
		University, 
		Milano, Piazza 
		dell'Ateneo Nuovo, 1, 20126 Milan, Italy, 
		\href{mailto:marco.repetto@unimib.it}{\nolinkurl{marco.repetto@unimib.it}}}
		 ~\footnote{Digital Industries, Siemens Italy, Via Privata Vipiteno 4, 20128 Milan, Italy,
		\href{mailto:marco.repetto@siemens.com}{\nolinkurl{marco.repetto@siemens.com}}}}
\date{\today}

\begin{document}
	\maketitle
	\begin{abstract}	
	Academic research and the financial industry have recently paid great 
	attention to Machine Learning algorithms due to their power to solve 
	complex learning tasks. In the field of firms' default prediction, 
	however, the lack of interpretability has prevented the extensive 
	adoption 
	of the black-box type of models. To overcome this drawback and maintain 
	the 
	high performances of black-boxes, this paper relies on a model-agnostic 
	approach. Accumulated Local Effects and Shapley values are used to 
	shape the predictors’ impact on the likelihood of default and rank them 
	according to their contribution to the model outcome. Prediction is 
	achieved by two Machine Learning algorithms (eXtreme Gradient Boosting 
	and FeedForward Neural Network) compared with three standard 
	discriminant models. Results show that our analysis of the Italian 
	Small and Medium Enterprises manufacturing industry benefits from the 
	overall highest classification power by the eXtreme Gradient Boosting 
	algorithm without giving up a rich interpretation framework.

		\par
		
		\textbf{Keywords:} Risk Analysis, Machine Learning, Intepretability, Small and Medium Sized Enterprises, Default Prediction.		
	\end{abstract}
	
	\setstretch{1.5}
	\hypertarget{introduction}{%
		\section{Introduction}\label{introduction}}
    The economy of the European Union (EU) is deeply grounded into Small and 
	Medium 
	Enterprises (SMEs).  SMEs represent about 99.8\% of the 
	active enterprises in the EU-28 non-financial business sector (NFBS), 
	accounting for almost 60\% of value-added within the NFBS and fostering the 
	workforce of the EU with two out 
	of every three jobs \citep{EU2019}. 
	
	Thus, there is a wide literature covering various economic 
	aspects of SMEs, with a particular attention to default prediction 
	\citep[for an up-to-date review see][]{ciampi2021rethinking}, which is of 
	interest not only for scholars but also for practitioners such as financial 
	intermediaries and for policy makers in their effort to support SMEs and to 
	ease credit constraints to which they are naturally exposed 
	\citep{andries2018financial,cornille2019heterogeneous}. 

Whether it is for private credit-risk assessment or for public funding, 
independently on the type of data imputed to measure the health status of a 
firm, 
prediction of default should success in two aspects: maximise correct 
classification and clarify the role of the variables involved in the 
process. Most of the times, the contributions based on 
Machine 
	Learning (ML) techniques neglect the latter aspect, being rather focused on 
	the former, often with better results with respect to parametric 
	techniques that provide, on the contrary, a clear framework for 
	interpretation. In other words ML techniques rarely deal with 
	\textit{interpretability} which, according to a recent document released by 
	the European Commission, should 
	be kept "in mind from the start" \citep{EC2019}.

Interpretability is central when a model is to be 
applied in the practice, both in terms of managerial choices and compliance: it 
is a fundamental requisite to bring a model into production.
Interpretable models allow risk managers and decision-makers to understand its 
validity by looking at its inner workings and helping in the calibration 
process. The European Commission itself encourages organizations to build 
	trustworthy Artificial Intelligence (AI) systems (including ML techniques) 
	around 
	several pillars: one of them is transparency, which encompasses  
	traceability, explainability and open communication about the limitations 
	of the AI system \citep{EC2020}.  		

Consequently, ML models -no matter how good in classifying default- should 
be made readable or their inherent uninterpretable nature will prevent their 
spreading in the literature on firms' default prediction as well as their use 
in contexts regulated by transparency norms. 
	
This work try to fill this gap by applying two different kind of ML models 
(FeedForward Neural Network \citep{haykin1999} and eXtreme Gradient Boosting 
\citep{chen2016} to Italian 
Manufacturing SMEs' 
default prediction, with a special attention to interpretability. Italy
represents an ideal testing ground for SMEs default prediction since its 
economic framework is even more shaped by firms up to this size with 
respect to the average EU country \citep{EU2019}. Default was assessed on the 
basis of 
	the firms' accounting information 
	retrieved from Orbis, a Bureau van Dijk (BvD) dataset.
	
	The main original contribution of the paper is to address ML models' 
	interpretability into the context of default prediction.  
	Our approach is based on model 
		agnostic-techniques and adds Accumulated Local Effects \citep[ALEs,][]{ 
		apley2020visualizing} to the  
		Shapley 
		values \citep[already applied in][]{bussmann2021explainable}. Using 
		these techniques we can rank 
	the variables in terms of 
	their contribution to the classification and determine their impact on 
	default 
	prediction, characterizing risky thresholds and non-linear patterns. 
	Robustness of the ML models hyperparameters was taken care of by Montecarlo 
	Cross-Validation and 
	substantial class imbalance between defaulted and survived firms was 
	reduced  
	through undersampling of the latter into the cross-validation training 
	sets. Another contribution of the paper is the benchmarking of the ML 
	models' outcome with  Logistic, Probit and with Binary Generalized
	Extreme Value Additive (BGEVA) classifications, both 
	according to standard performances metrics and to the role played by the 
	different variables. We also supply ALEs for the parametric models, in 
	order to get additional insights with respect to variables significance and 
	to guarantee a common ground for comparison.  
	
	We obtain at least three interesting results. First, thanks to our research 
	design, all the 
	proposed models 
		work fairly well according to all the metrics. Second, eXtreme Gradient Boosting (XGBoost) 
		outperformed the others mainly for total 
		classification accuracy and default prediction rate. Third, the impact 
		of the variables retrieved from XGBoost is fully consistent with 
	the economic literature, whereas the same cannot be said for the competing 
	models.

	The remainder of the paper is organized as follows.
	\protect\hyperlink{literature}{Section 2} gives an overview of the 
	(necessarily) recent literature concerning ML intepretability.
	\protect\hyperlink{data}{Section 3} provides a description of the dataset 
	and of the features we use throughout the modelling.
	\protect\hyperlink{methodology}{Section 4} discusses our methodology, 
	briefly reviewing the models fundamentals, the techniques employed 
	for interpretability and the research design. 
	\protect\hyperlink{results}{Section 5} presents the results and discusses 
	the most relevant findings. \protect\hyperlink{conclusions}{Section 6} 
	concludes.
	
	\hypertarget{literature}{%
		\section{Literature review}\label{literature}}
The ability to predict corporate failure has been largely 
	investigated in credit risk literature since the 1970s 
	\citep{laitinen1999early}. On the one hand, the academic interest in the 
	topic has grown after the global financial crisis (2007-2009) that has 
	highlighted the need for more consistent models to assess the risk of 
	failure of SMEs \citep{oliveira2017integrating} and it is renewed today 
	with the current pandemic impact on the companies of all sizes 	
	\citep{ciampi2021rethinking}. On the other hand, a good part of the 
	financial industry has shown great attention to statistical algorithms 
	that prioritize the pursuit of 
	predictive power. Such a trend has been registered by recent surveys,  
	showing  
	that credit institutions are gradually embracing ML 
	techniques 
	in different areas of credit risk management, credit scoring and monitoring 
	\citep{BoEn19, alonso2020machine, IFF20}. Among all, the biggest annual 
	growth in the adoption of highly performing algorithms has been observed in 
	the SMEs sector \citep{IFF19}.

For these reasons, new modeling techniques have been 
successfully employed in predicting SMEs default, including Deep 
Learning \citep{mai2019deep}, boosting 
\citep{moscatelli2020corporate, kou2021bankruptcy}, Support Vector Machine 
\citep{kim2010support, gordini2014genetic, zhang2015credit}, Neural Networks 
\citep{alfaro2008bankruptcy, ciampi2013small, geng2015prediction} and 
two-stage classification \citep{du2016two}, to name a few.
However, they have been applied mainly in order to improve classification  
accuracy with respect to the standard linear models, rather than in terms of 
causality patterns. But the latter is no longer a negligible aspect, both  
for academic research and for management of regulated financial services: it 
has become overriding,  since the European Commission and other European 
Institutions released a number of regulatory standards on Machine Learning 
modeling. 

The Ethics Guidelines for Trustworthy AI \citep{EC2019} 
	and the Report on Big Data and Advanced Analytics \citep{EBA2020} 
	illustrate the principle of explicability of ML algorithms that have to be
	transparent and fully interpretable, including for those directly and 
	indirectly affected. Indeed, as the Commission points out, predictions, 
	even accurate, without explainability measures are not able to 
	foster responsible and sustainable AI innovation. The pillar of 
	transparency (fourth among seven), somewhat combines explainability and 
	interpretability 
 of a model, referring to interpretability as the "concept of 
	comprehensibility, explainability, or understandability" \citep{EC2020}. 

		The difference in meaning between interpretability and explainability, 
		synonymous in the dictionary, has been addressed by the recent ML 
		literature which recognizes the two words a conceptual distinction   
		related to 	different properties of the model and knowledge aspects  
	\citep{doran2017does, lipton2018mythos}. A clear indication about the 
	distinction is given by \cite{montavon2018methods} that defines 
	interpretation as a mapping of an abstract 
	concept into a domain that 
	the human expert can perceive and comprehend  and explanation as a 
	collection of features of the interpretable domain that have 
	contributed to a given example to produce a decision. Roughly speaking, 
	 interpretability is defined as the ability to spell out or to provide 
	the meaning in understandable terms to a human 
	\citep{doshi2017towards, guidotti2018survey}, whereas explainability is 
	identified as the capacity of revealing the causes underlying the decision 
	driven by a ML method \citep{arrieta2020explainable}.
	
There are several approaches to ML interpretability in the literature, 
classified in two main categories: ante-hoc and post-hoc methods. 
		Ante-hoc methods employ intrinsically interpretable models (e.g., 
		simple decision tree or regression models, also called white-box) 
		characterized by a simple 
		structure. They rely on model-specific interpretations depending on 
		examination of internal model parameters. Post-hoc methods 
		instead provide a reconstructed interpretation of decision records 
		produced by a black-box model in a reverse engineering logic 
		\citep{ribeiro2016model, du2019techniques}. These methods reckon on  
		model-agnostic interpretation where internal model parameters are not 
		inspected.
	
So far, ante-hoc approaches were widely used in the SMEs default prediction  
	literature that counts contributions employing mainly white-box models as 
	Logistic 
	regression \citep[see for 
	example][]{sohn2007random, lin2012predicting, modina2014default, ciampi2015corporate},
	 Survival analysis 
	\citep{holmes2010analysis, gupta2015forecasting, el2016effect, gupta2018empirical}
	 or Generalised Extreme Value regression 
	\citep{calabrese2016bankruptcy}. The empirical evidences and the variables' 
	effect on the outcome are interpreted in an inferential testing setting, so 
	that the impact of the predictors and the results' implications are always 
	clear to the reader.  
	An alternative strategy can be found in \cite{liberati2017advances} that interprets a kernel-based classifier for default prediction via a surrogate model, but the results and the variables impact are conditioned to the regression fit to the data.
	
	On the contrary, post-hoc methods have been rarely used in this field and 
	comprehend Partial Dependence (PD) plots 
	\citep{friedman2001greedy}, Local Interpretable Model-agnostic 
	Explanations (LIME) \citep{ribeiro2016should} and the  SHAP values 
	\citep{lundberg2017unified}, all of them providing detailed model-agnostic 
	interpretation of the 
	complex ML algorithms employed.  \cite{jones2019predicting}, 
	\cite{sigrist2019grabit} and \cite{jabeur2021catboost} used the PD to 
	identify the relevant variables' subset and to measure the change of 
	the average probability of default with respect to the single features. 
	\cite{stevenson2021value} and \cite{yildirim2021big}, instead, applied LIME 
	and SHAP values to rank the most important variables
	and to provide the features impacts on the output prediction respectively. 
	In this paper we employ the 
	Accumulated Local Effects (ALE), a model-agnostic technique that represents 
	a 
	suitable 
	alternative to PDs when the features are highly correlated, without 
	providing incoherent values \citep{apley2020visualizing}. Since ALE is a 
	newest approach, its usage is far more limited and it has not yet spread  
	in the bankruptcy prediction area. An isolate application to the recovery 
	rate forecasting of non-performing loans can be encountered in the credit 
	risk field \citep{bellotti2021forecasting}.

	\hypertarget{data}{%
		\section{Data Description}\label{data}}
	The data of this study are retrieved from BvD database
	, which 
	provides financial and accounting ratios from balance sheets of the 
	European private companies. We have restricted our focus on Italian 
	manufacturing SMEs for several reasons. 
	Italy is the second-largest manufacturing country in the EU 
	\citep{bellandi2020} and this sector generates more than 30\% of the 
	Italian GDP \citep{eurostat2018}. 
	Differently from SMEs in other EU countries, Italian SMEs trade 
	substantially more than large firms and manufacturing, in particular, 
	drives both import and exports 
	\citep{abel2018}. Moreover, according to \cite{ciampi2009} and 
	\cite{ciampi2013small}, predictive models have better performances when 
	trained for a specific sector as this avoids pooling heterogeneous firms.

	To define our sample, we filtered the database 
	both by country and NACE codes (from 10 to 33) and we employed the European 
	Commission definition \citep{eu2003} of Small 
	and Medium Enterprises. We retrieved only firms with 
	annual turnover of fewer than 50 million euros, the number of employees 
	lower than 250 and a balance sheet of fewer than 43 million euros.  Among 
	those, we classified as defaults all the enterprises that entered 
	bankruptcy or a liquidation procedure, as well as active companies that 
	had not repaid debt (default of payment), active companies in 
	administration or receivership or under a scheme of the arrangement, 
	(insolvency proceedings), which in Orbis are also considered in default.  
	Consistently with 
	the literature, we excluded dissolved firms that no longer exist as a legal 
	entity when the reason for dissolution is not specified \citep{altman2007modelling, altman2010sabato, andreeva2016comparative}. 
	This category in fact encompasses firms that may not necessarily experience 
	financial difficulties. The resulting dataset contains 105,058 SMEs with a 
	proportion of 1.72\% (1,807) failed companies.
	
	 The accounting indicators, which refer to 2016 for predicting the status 
	 of the firms in 2017, have been selected among the most frequently used in 
	 the SMEs default literature and are the following:
	\begin{itemize}
		\item
		Cash flow: computed as net income plus depreciations \citep{lin2012predicting, calabrese2013modelling, ciampi2013small, succurro2014impact, michala2013, jones2015, andreeva2016comparative};
		\item
		Gearing ratio: computed as the ratio between total debt and total 
		assets \citep{altman2007modelling, ciampi2009, fantazzini2009random, altman2010sabato, lin2012predicting, jones2015, andreeva2016comparative, psillaki2010};
		\item
		Number of employees, as a size measure from an input perspective 
		\citep{andreeva2016comparative};
		\item
		Profit margin: measured as profit/loss before tax over the operating 
		revenue \citep{ciampi2009, altman2010sabato, lin2012predicting, ciampi2013small, andreeva2016comparative};
		\item
		ROCE: computed as profit/loss before tax over capital employed, which 
		is given as total assets minus current liabilities \citep{lin2012predicting, andreeva2016comparative, nehrebecka2018};
		\item
		ROE: computed as profit/loss before tax over shareholders' funds \citep{ciampi2009, altman2010sabato, ciampi2013small, succurro2014impact, jones2015, andreeva2016comparative};
		\item
		Sales: in thousands Euro, measuring the output side of firm size \citep{succurro2014impact};        
		\item
		Solvency ratio: computed as shareholders' funds over total assets \citep{ciampi2009, calabrese2013modelling, jones2015, andreeva2016comparative, calabrese2016bankruptcy, nehrebecka2018};
		\item
		Total assets: in thousands Euro, as a measure of total firm resources  
		\citep{lin2012predicting, michala2013, andreeva2016comparative}.
	\end{itemize}
	
	As a quick preview of the expected relationship between the single 
	predictors and the likelihood of default, we have computed the average 
	values and standard deviations of the variables separately for survived and 
	defaulted 
	firms (see table \ref{tab:summaryvar}). 
	\begin{table}[ht!]
		
		\caption{\label{tab:summaryvar}Summary statistics by survived and 
			failed firms}
		\centering
		\footnotesize
		\begin{tabular}[t]{lrrrr}
			\toprule
			\multicolumn{1}{c}{ } & \multicolumn{2}{c}{Survived} & 
			\multicolumn{2}{c}{Failed} \\
			\cmidrule(l{3pt}r{3pt}){2-3} \cmidrule(l{3pt}r{3pt}){4-5}
			Variable & Mean & St. dev. & Mean & St. dev. \\
			\midrule
			Cash flow			& 236,802 	& 934,877 	& -278,521 	& 1.636,028 
			\\
			Gearing ratio 		& 24,807 	& 23,093	& 22,166 	& 26,01 \\
			Number of employees & 16,506 	& 24,385 	& 11,08 	& 19,531 \\
			Profit margin 		& -2,736 	& 610,488 	& -106,845 	& 2.190,012 
			\\
			ROCE 				& 12,335 	& 516,765 	& 66,367	& 2.284,001 
			\\
			ROE 				& 23,02 	& 314,135 	& 7,146 	& 971,112 \\
			Sales 				& 3.427,163 & 6.301,229 & 1.259,695 & 2.940,01 
			\\
			Solvency ratio		& 27,101 	& 24,315 	& -1,044	& 37,342 \\
			Total assets 		& 3.904,129 & 12.098,09 & 1.921,689 & 5.149,559 
			\\
			\bottomrule
		\end{tabular}
	\end{table}

	In line with \cite{andreeva2016comparative},
	we can see on average weakest liquidity, smallest size and deficient 
	leverage for defaulted
	firms. The 
	Profit margin and ROE are highest for surviving firms, whereas the 
	remaining 
	profitability index, ROCE, shows a larger mean among defaulted 
	firms. ROCE should be, according to common sense, negatively 
	related to default. However, some studies found its impact non-significant 
	coherently with 
	the low-equity dependency of small businesses \citep{ahelegbey2019, 
	giudici2020}, while others attest a positive effect of ROCE on default 
	with a caveat for large values \citep{calabrese2016bankruptcy}.

\hypertarget{methodology}{%
\section{Methodology}\label{methodology}}

\hypertarget{models-employed}{%
	\subsection{White-box versus black-box models}\label{models-employed}}

The models we apply can be broadly classified as white-box, or interpretable,  
and black-box but post-hoc interpretable in the model-agnostic framework. 

In the first category, Logistic Regression (LR) and Probit were selected among 
the most recurrent models in the economics literature, where the accent on the 
factors 
impacting default is certainly of primary importance. These 
models frequently serve as a 
benchmark for classification when a new method is proposed. The third model, 
BGEVA \citep{calabrese2013modelling}, comes from the Operational 
Research literature and is based on the quantile function of a Generalized 
Extreme Value random variable. The main strengths of BGEVA  are robustness, 
accounting for  
non-linearities and preserving interpretability. 

The black-box models we use are XGBoost and FeedForward Neural Network (FANN). These models are by nature uninterpretable since the explanatory variables pass multiple trees (XGBoost)
or layers (FANN), thus generating an output for which an understandable explanation cannot be provided. 

The XGBoost algorithm was found to provide the best performance in default 
prediction with respect to LR, Linear Discriminant Analysis, and Artificial Neural Networks \citep{petropoulos2019, bussmann2021explainable}.
The algorithm builds a sequence of shallow decision trees, which are trees with 
few leaves. Considering a single tree one would get an 
interpretable model taking the following functional form:

\begin{equation}
	f(x) = \sum^M_{m=1} \theta_m I(x \in R_m)
\end{equation}

where \(M\) covers the whole input space with \(R_1, ... R_M\) non-overlapping 
partitions, \(I(\cdot)\) is the indicator function, and \(\theta_m\) is the 
coefficient associated with partition $R_m$.
In this layout, each subsequent tree learns from the previous one and improves  
the prediction \citep{friedman2001greedy}.

As a competing black-box model we chose the FANN, which is widely used and well performing in SMEs' default prediction \citep{ciampi2021rethinking} and in several works on retail credit risk modeling \citep{west2000, baesens2003, west2005}.
FANN consists of a direct acyclic network of nodes organized in densely 
connected layers, where inputs, after been weighted and shifted by a bias term, 
are fed into the node's activation function and influence each subsequent layer 
until the final output layer.
In a simple classification task, in which only one hidden layer is necessary, a 
FANN can be described by:

\begin{equation}
	\label{FANN_eqn}
	P(y=1|x) = \sum^M_{m=1} w^1_m a(w^0_m x + b_m)
\end{equation}

where the weighted combination of the inputs is shifted by the bias \(b\) and 
activated by the function \(a\). Note that imposing \(a(\cdot)\) as the sigmoid 
function in equation \ref{FANN_eqn} the model collapses to a LR.

\hypertarget{model-agnostic-interpretability-techniques}{%
\subsection{Model-agnostic 
interpretability}\label{model-agnostic-interpretability-techniques}}

To achieve the goal of interpretability we make use of two different and 
complementary model-agnostic techniques. 
First, we use the global Shapley Values \citep{shapley1953} to provide 
comparable information on the single feature contributions to the model output. 
Global Shapley Values have been already proposed in the SMEs default prediction 
literature by \cite{bussmann2021explainable}.

However, global Shapley Values do not provide any information about the direction and shape of the variable effects. To get this information, we resort to ALEs \citep{apley2020visualizing}. 
ALEs, contrary to Shapley Values, offer a visualization of the path according to which the single variables impact on the estimated probability of default. 

To further clarify the improvement that ALEs bring to interpretability in our setting, we briefly contextualize the method and sketch its fundamentals. 

The first model-agnostic approach for ML models' interpretation to appear in 
the literature was Partial Dependence (PD), proposed by \cite{friedman1991} in the early '90s. 
PD plots evaluate the change in the average predicted value as specified 
features vary over their marginal distribution \citep{goldstein2015}.
In other words, they measure the dependence of the outcome on a single feature when all of the others are marginalized out. 
Since their first formulation, PD plots have been used extensively in many 
fields but seldom in the credit risk literature, whit a recent application by  
\cite{su2019}.

One of the main criticisms moved to PD is on its managing the relationships within features. 
The evaluation of PD on all the possible feature configurations carries the risk of computing points outside the data envelope: such points, intrinsically artificial, can result in a misleading effect of some features when working on real datasets.

Due to this fallacy, and because of the renewed interest in complex deep learning models as Artificial Neural Networks, many new methodologies have been proposed. 
With Average Marginal Effects (AMEs), \cite{hechtlinger2016} suggested to 
condition the PD to specified values of the data envelope. 
\cite{ribeiro2016should} went the opposite direction presenting a local approximation of the model through simpler linear models, the so-called Local Interpretable Model-agnostic Explanations (LIME). 
In subsequent research, they also worked on rule-based local explanations of 
complex black-box models \citep{ribeiro2018}. 
Shapley Additive exPlanations (SHAP) was introduced by  
\cite{lundberg2017unified} to provide a human understandable and local 
Shapley evaluation.

In this framework, ALEs constitute a further refinement of both PD and AMEs.
 They avoid the PD plots-drawback of assessing 
variables' 
effects outside the data envelope, 
generally occurring when features are highly correlated 
\citep{apley2020visualizing}, as in the case of many accounting 
indicators \citep{ciampi2009, altman2010sabato, ciampi2015corporate}. 
Furthermore, ALEs do not simply condition on specified values of the 
data envelope as AMEs do, but take first-order differences conditional on the 
feature space partitioning, eventually eliminating possible bias derived from 
features' relationships.

Specifically, computing the ALE 
implies the evaluation of the following type of 
function:

\begin{equation}
	\label{ALE_eq}
	ALE_{\hat{f}, S} (x) = \int_{z_{0,S}}^x \left[ \int \frac{\delta 
	\hat{f}(z_S, X_{ \setminus S})}{\delta z_S} d\mathcal{P}(X_{\setminus 
	S}|z_S)\right] dz_S - constant
\end{equation}

where:

\begin{itemize}
	\tightlist
	\item
	\(\hat{f}\) is the black-box model;
	\item
	\(S\) is the subset of variables' index;
	\item
	\(X\) is the matrix containing all the variables;
	\item
	\(x\) is the vector containing the feature values per observation;
	\item
	\(z\) identifies the boundaries of the K partitions, such that \(z_{0,S} = 
	min(x_S)\).
\end{itemize}

The expression in equation \ref{ALE_eq} is in principle not 
model-agnostic as it requires accessing the gradient of the model:
\(\nabla_{z_S}\hat{f} = \frac{\delta \hat{f}(z_S, X_{ \setminus S})}{\delta 
z_S}\) but this is not known or even not existent in certain 
black-boxes.
As a replacement, finite differences are taken to the boundaries of 
the partitions, \(z_{k-1}\) and \(z_{k}\).

Hence, the resulting formula to evaluate ALEs is: 

\begin{equation}
	\label{ALE_est}
	ALE_{\hat{f}, S} (x_S) = \sum_{k=1}^{k_S(x)}\frac{1}{n_S(k)} 
	\sum_{i:x_S^{(i)} \in N_S (k)} \left[\hat{f}(z_{k,j}, x_{\setminus 
	S}^{(i)}) - \hat{f}(z_{k-1,j}, x_{\setminus S}^{(i)}) \right] - \frac{1}{n} 
	\sum_{i=1}^n ALE_{\hat{f}, S} (x_S^{(i)})
\end{equation}

The replacement of the constant term in equation \ref{ALE_eq} by \(- 
\frac{1}{n} \sum_{i=1}^n ALE_{\hat{f}, S} (x_S^{(i)})\) in equation 
\ref{ALE_est} centers the plot, which is something missing in PD.
This makes it clear that, by evaluating predictions' finite differences conditional on \(S\) and integrating the derivative over features \(S\), ALEs disentangles the interaction between covariates. 
This way the main disadvantage of PD is solved. 

\hypertarget{research-design}{%
\subsection{Research design}\label{research-design}}
Our research design has been carried out according to \cite{lessmann2015}. 
We split the initial dataset into  training (70\%) and test  (30\%) sets \citep{james2013}. 
Then, through the Monte Carlo Cross-Validation procedure \citep{xu2001}, we 
estimate the models parameters and validate the estimated rules. 
More in detail, at each iteration we create a sub-training set and a validation 
set via random sampling without replacement so that the models learn from the 
training set whereas the assessment, based on performance metrics, is done on 
the validation set. 
This way, we also tune the hyperparameters of the algorithms when necessary. 

The training set serves as well to compute the Shapley values, based on the 
optimal rule, and to calculate the 
ALEs with corresponding bootstrap non-parametric confidence intervals 
\citep{davison2002, apley2020visualizing}.
Finally, we evaluated the models’ performance on the hold-out sample (test set).

We took into account also the severe unbalance in favour of survived firms to avoid over-classification of the majority class \citep{baesens2021}. 
In the learning phase, we employed perfectly balanced samples, obtained through random under-sampling of the survived firms.  
This sampling scheme combined the best classification 
performances with a drastic reduction in the computation time. Obviously the 
undersampling scheme was applied only to the training data, to avoid 
over-optimistic performance metrics on either the validation or the test set 
\citep{gong2017, santos2018}.

\hypertarget{results}{%
\section{Results}\label{results}}

The results are organized according to the performance and 
interpretation of the five models. 

The performance is measured by the proportion of 
failed and survived firms correctly identified (sensitivity and specificity) as 
well as by two global perfomance metrics: the Area Under the Receiver 
Operating Curve (AUC) and the H-measure (see 
Table  \ref{tab:modelresults}). 
The need for an additional indicator of global performance other than the AUC 
is motivated to assure robustness of the results. 
The AUC has indeed several advantages in providing consistent classifier 
performance \citep{lin2012predicting} and is widely adopted by practitioners 
\citep{lessmann2015}, but it suffers from using different misclassification 
cost 
distributions for different classifiers as outlined by  
\cite{hand2009} and \cite{hand2013}.
Thus, we add the H-measure that normalizes classifiers' cost distribution based on the expected minimum misclassification loss. 

Second, we cross-compare the role and weight of the variables 
among models and contextualize the results within the literature. 
The post-hoc interpretation of the black-box models is based on the Shapley 
values and ALEs. We report the ALEs also for interpretable models to exploit a 
common basis for predictors comparison without incurring in the 
"p-value arbitrage" when evaluating white-box models via p-values and 
 ML models via other criteria \citep{breeden2020survey}.

\hypertarget{overview}{%
\subsection{Performance}\label{overview}}

All competing models offer fair correct classification rates, but the ones that 
score globally best 
are black-box models, in terms of both AUC and H-measure metrics. The FANN 
reaches the 
highest H-measure and specificity while it's last as far as correct 
classification of default is concerned (with a sensitivity not reaching 70\%, 
see Table 
\ref{tab:modelresults}). On the contrary, 
the XGBoost algorithm provides the best 
 default prediction (showing, by far, the largest sensitivity) with a 
 reasonable  
 classification of 
 survivors, resulting in the highest AUC and the second-best 
 H-measure.

\begin{table}[!ht]
	\caption{\label{tab:modelresults} Models' performances on the test set.}
	\centering
	\begin{tabular}[t]{lllll}
		\toprule
		Model & Sensitivity & Specificity & H-measure & AUC\\
		\midrule
		Feedforward Artificial Neural Network & 0.694 & \textbf{0.829} & 
		\textbf{0.391} & 0.827\\
		eXtreme Gradient Boosting & \textbf{0.821} & 0.719 & 0.383 & 
		\textbf{0.843}\\
		BGEVA Model & 0.752 & 0.727 & 0.331 & 0.819\\
		Logistic Regression & 0.745 & 0.736 & 0.303 & 0.809\\
		Probit Model & 0.738 & 0.737 & 0.299 & 0.809\\
		\bottomrule
	\end{tabular}
\end{table}

 The interpretable models are ranked consistently by the two global measures of 
 performance in the following order: BGEVA, LR and Probit.  
These results confirm the trade-off 
between  performance and ante-hoc interpretability highlighted in the 
retail credit risk modeling literature \citep{baesens2003, lessmann2015} and in 
previous works on Italian SMEs \citep{ciampi2013small}.

All in all, undersampling the training set has a balancing effect on the rate 
of correct 
prediction for either class. This improves  
global classification not only through FANN, but also when applying Logistic 
Regression, as compared for instance with the results on the same kind of 
variables of 
\cite{ciampi2013small} or \cite{modina2014default}, the latter for both 
 techniques.

\hypertarget{interpreting-the-models}{%
\subsection{Interpretation}\label{interpreting-the-models}}

Most of the variables  have 
non-significant effects on the probabilities of default estimated by 
white-box 
models, as long as these effects are ascertained by p-values (table 
\ref{tab:interpretable}). Three variables display a 
significant and non-null coefficient, no matter the 
model: Sales\footnote{In the text we refer to Sales, Total Assets 
	and Number of 
Employees for readability reasons. However, we have transformed them through 
logarithms as common in the literature \citep{altman2007modelling, altman2010sabato, psillaki2010}}, the Solvency Ratio and the Cash 
flow, all with an adverse effect on the probability of default. 

The negative impact exerted by Sales on default, recurrent in 
many works, is not surprising since Sales is one of the main proxies of a 
company's 
size and it is well known that largest firms tend to overcome demand shocks 
better than the smaller firms \citep{psillaki2010, ciampi2015corporate}. This 
is also consistent with the means reported in Table \ref{tab:summaryvar} for 
the two groups of firms. Apparently, the size effect is captured exclusively by 
the output-side variable since the other size proxies, the Number of employees 
and the Total Assets, both highly correlated with Sales 
\citep{jabeur2021catboost}, do not have instead significant effects. 

As expected, firms with a strongest leverage (Solvency ratio)  and higher 
liquidity (Cash flow) are less likely to default \citep{ciampi2009, calabrese2013modelling, michala2013, andreeva2016comparative}.  

\begin{figure}
	\centering
	\subfloat[a][]{
		\includegraphics[width=0.80\textwidth]{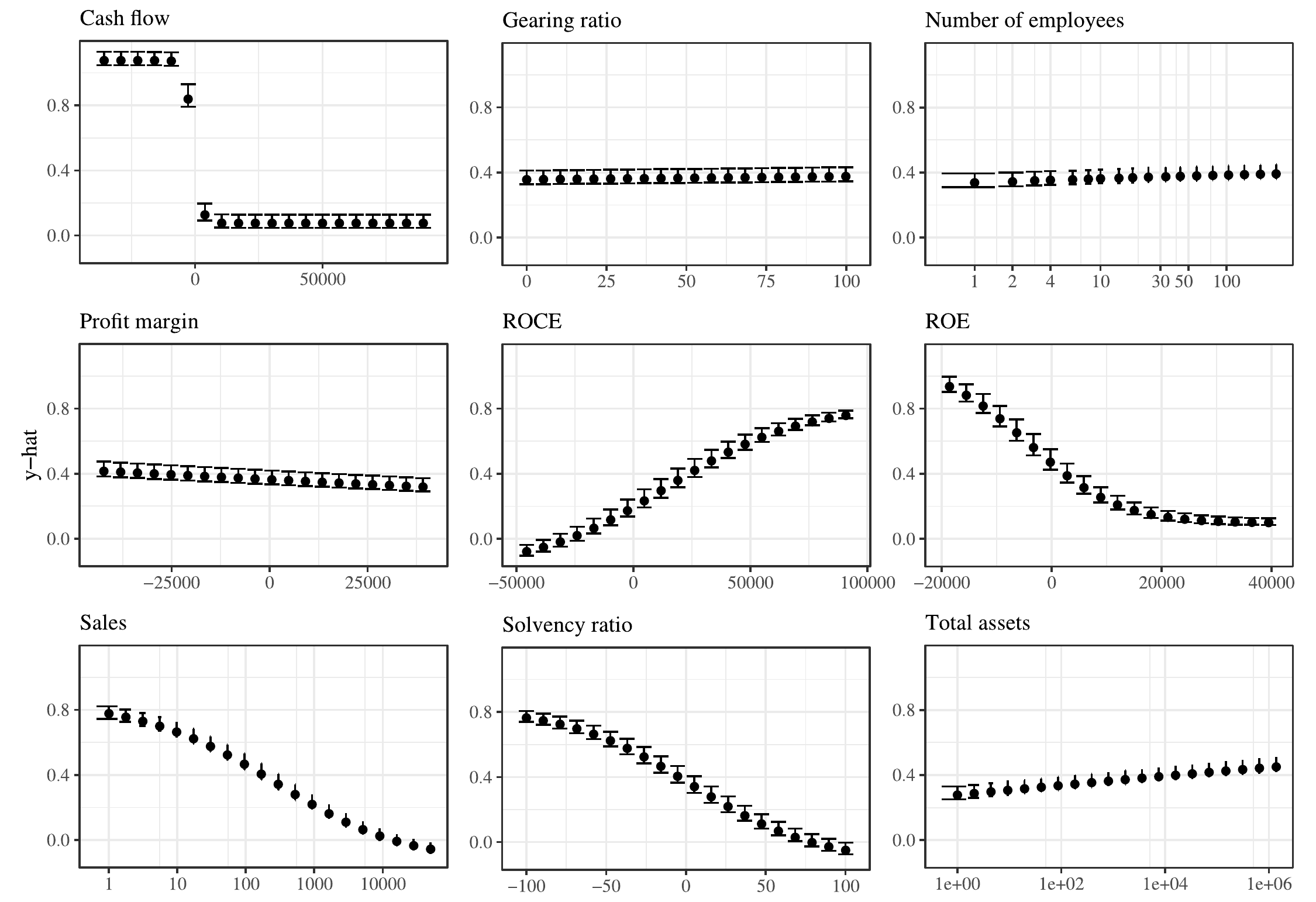}
	} \\
	\subfloat[b][]{
		\includegraphics[width=0.80\textwidth]{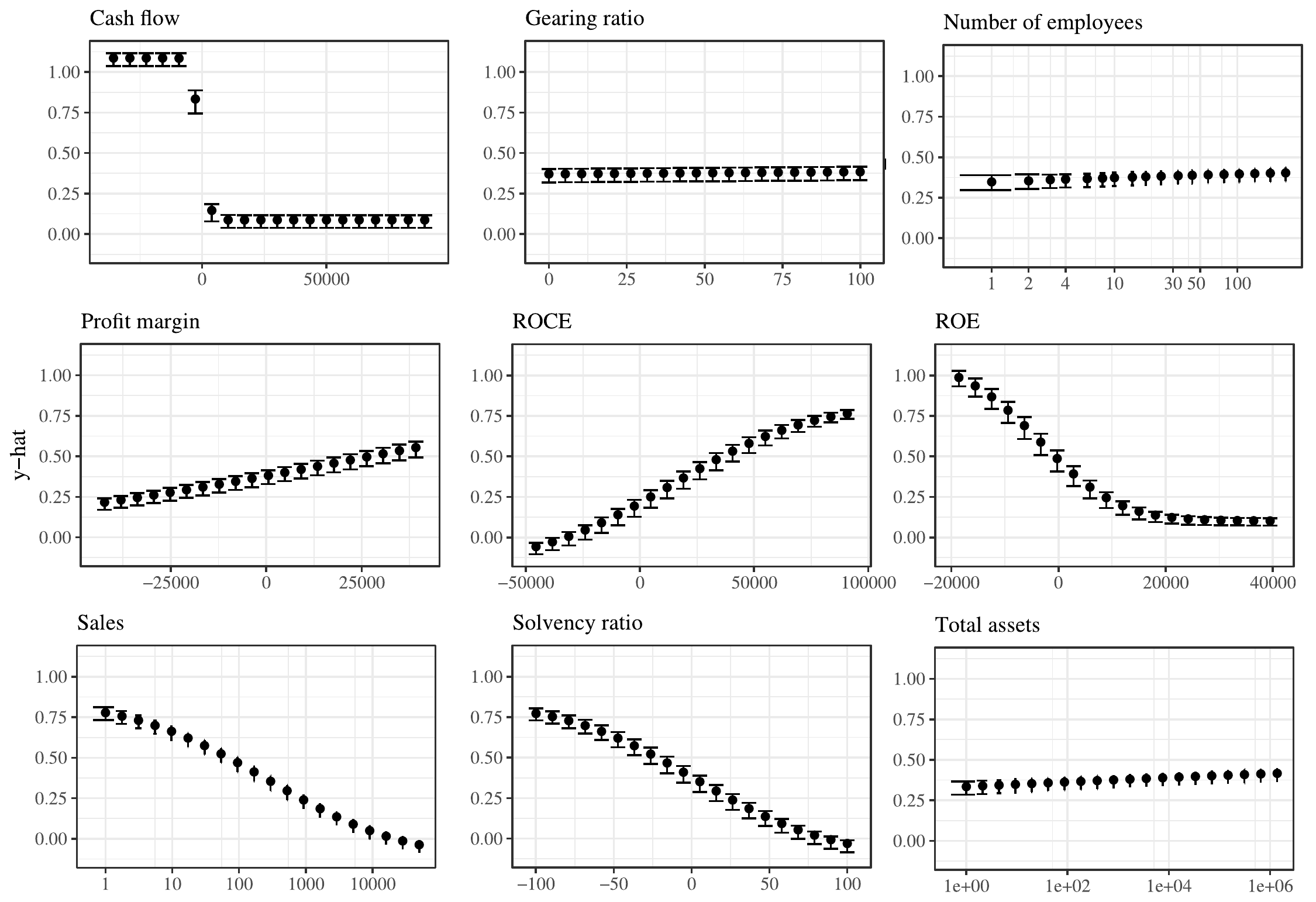}
	}
	\caption{
		Accumulated Local Effects of the LR (a) and Probit (b) models with 
		superimposed bootstrap 5\%-95\% confidence intervals.
		The ALEs for Sales, Total Assets and Number of Employees are calculated on log-transformation of the variables but depicted on anti-log values to enhance readability.}\label{fig:alexprobit}
\end{figure}

\begin{figure}[!ht]
			\begin{center}{ 
		\includegraphics[width=0.80\textwidth]{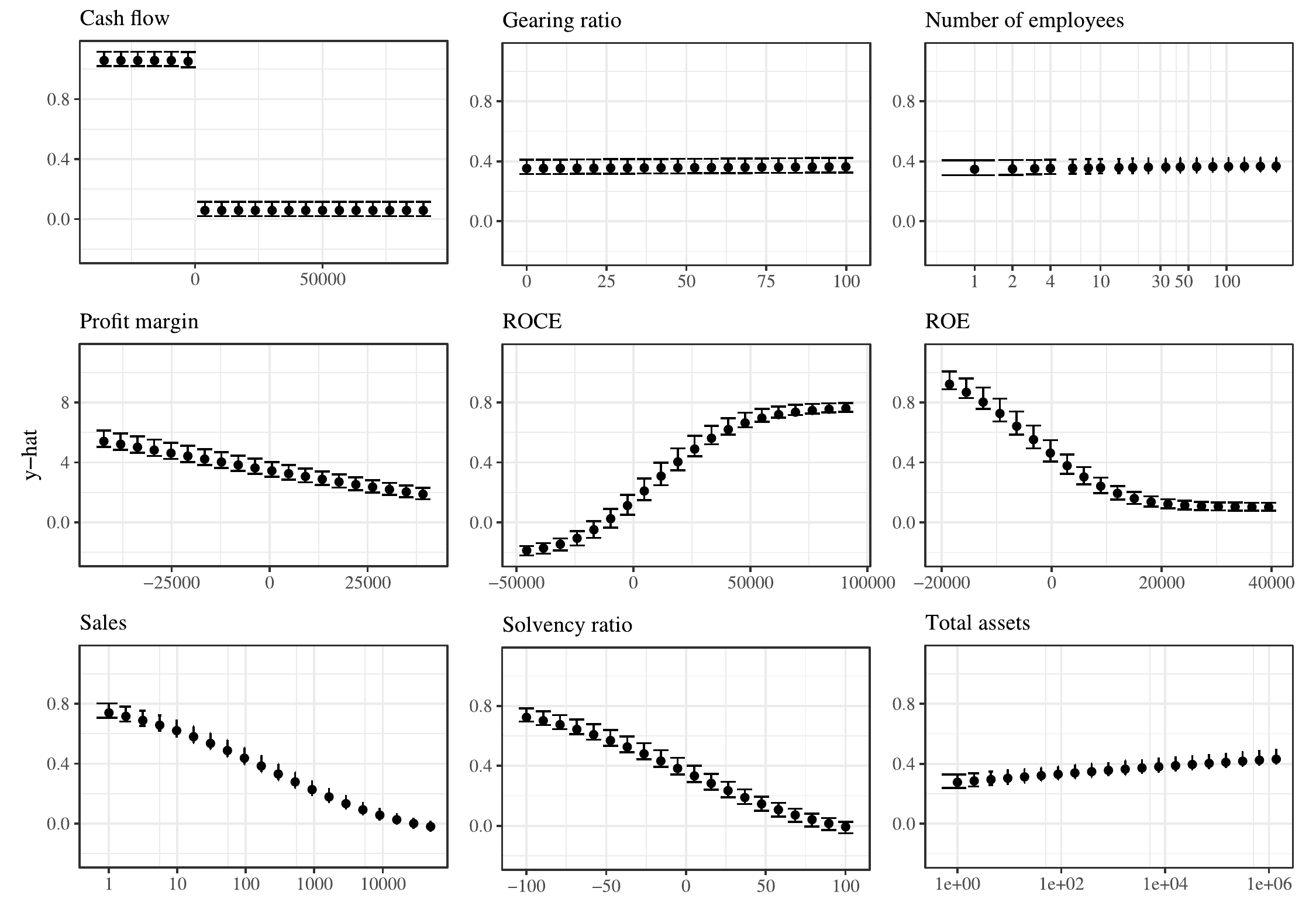}
		 
		}\end{center}
			\caption{Accumulated Local Effects of the BGEVA model with  
				superimposed  bootstrap 5\%-95\% confidence intervals.  The 
				ALEs for Sales, Total 
				Assets and Number of Employees are calculated on 
				log-transformation of the 
				variables but depicted on anti-log values to enhance 
				readability.}\label{fig:alexbgeva}
	\end{figure}

Notice that profitability measures, rather unexpectedly, do not impact on the 
probability of default according to significance criteria. BGEVA signals a 
significant ROCE but the estimated coefficient is zero. To gain 
additional  
insights, we can turn to the ALEs: the three common significant variables can 
be interpreted likewise since they all follow a non-flat path. However, while 
the models' coefficients for the Solvency ratio and Cash flow describe almost 
neutral effects on the outcome (with an odds-ratio of 1 for the Cash flow in 
the Probit model, see Table \ref{tab:interpretable}), post-hoc interpretation 
reveals a marked decreasing effect for the former and a clear non-linear 
pattern for the latter.  On the other hand, and 
contrary to the p-value reading, we can observe that Profit margin and 
ROE do reduce the 
probability of default, whereas ROCE increases it according to the LR, Probit 
(see figure \ref{fig:alexprobit}, panels (a) and (b) respectively) and to the 
BGEVA model (figure 
\ref{fig:alexbgeva}).

\begin{landscape}
	\begin{table}
		\caption{\label{tab:interpretable} Estimates and summary statistics for 
			the 
			Probit, Logistic Regression, and BGEVA models on the test set 
			(Significant 
			variables in bold).}
		\centering
		\begin{tabular}[t]{lrrrrrrrrr}
			\toprule
			\multicolumn{1}{c}{ } & \multicolumn{3}{c}{Probit Model} & 
			\multicolumn{3}{c}{Logistic Regression} & \multicolumn{3}{c}{BGEVA 
				Model} \\
			\cmidrule(l{3pt}r{3pt}){2-4} \cmidrule(l{3pt}r{3pt}){5-7} 
			\cmidrule(l{3pt}r{3pt}){8-10}
			& Odds ratio & Std. error & p-value & Odds ratio & Std error & 
			p-value & Estimate & Std. error & p-value\\
			\midrule
			(Intercept) & 6.195 & 0.134 & 0.000 & 21.256 & 0.233 & 0.000 & 
			2.087 & 0.137 & 0.000\\
			Cash flow & \textbf{1.000} & 0.000 & \textbf{0.000} & \textbf{0.999} & 0.000 & 
			\textbf{0.000} & \textbf{-0.001} & 0.000 & \textbf{0.000}\\
			Gearing ratio & 1.000 & 0.001 & 0.713 & 1.001 & 0.002 & 0.594 & 
			0.000 & 0.001 & 0.756\\
			Number of employees & 1.081 & 0.078 & 0.319 & 1.135 & 0.131 & 0.332 
			& 0.033 & 0.081 & 0.683\\
			Profit margin & 1.000 & 0.000 & 0.535 & 1.000 & 0.000 & 0.947 & 
			0.000 & 0.000 & 0.800\\
			ROCE & 1.000 & 0.000 & 0.256 & 1.000 & 0.000 & 0.302 & \textbf{0.000} & 
			0.000 & \textbf{0.027}\\
			ROE & 1.000 & 0.000 & 0.240 & 1.000 & 0.000 & 0.275 & 0.000 & 0.000 
			& 0.285\\
			Sales & \textbf{0.526} & 0.066 & \textbf{0.000} & \textbf{0.316} & 0.120 & 
			\textbf{0.000} & \textbf{-0.637} & 0.064 & \textbf{0.000}\\
			Solvency ratio & \textbf{0.985} & 0.001 & \textbf{0.000} & \textbf{0.973} & 0.002 & 
			\textbf{0.000} & \textbf{-0.015} & 0.001 & \textbf{0.000}\\
			Total assets & 1.044 & 0.064 & 0.503 & 1.166 & 0.112 & 0.172 & 
			0.090 & 0.066 & 0.174\\
			\bottomrule
		\end{tabular}
	\end{table}
\end{landscape}

 Another counterintuitive effect is revealed by the ALEs plot of the Profit 
 margin for the Probit (figure \ref{fig:alexprobit}, panel (b)), which could 
 partially explain the suboptimal classification performance of the same model.

\begin{figure}[!ht]
	\centering
	\includegraphics{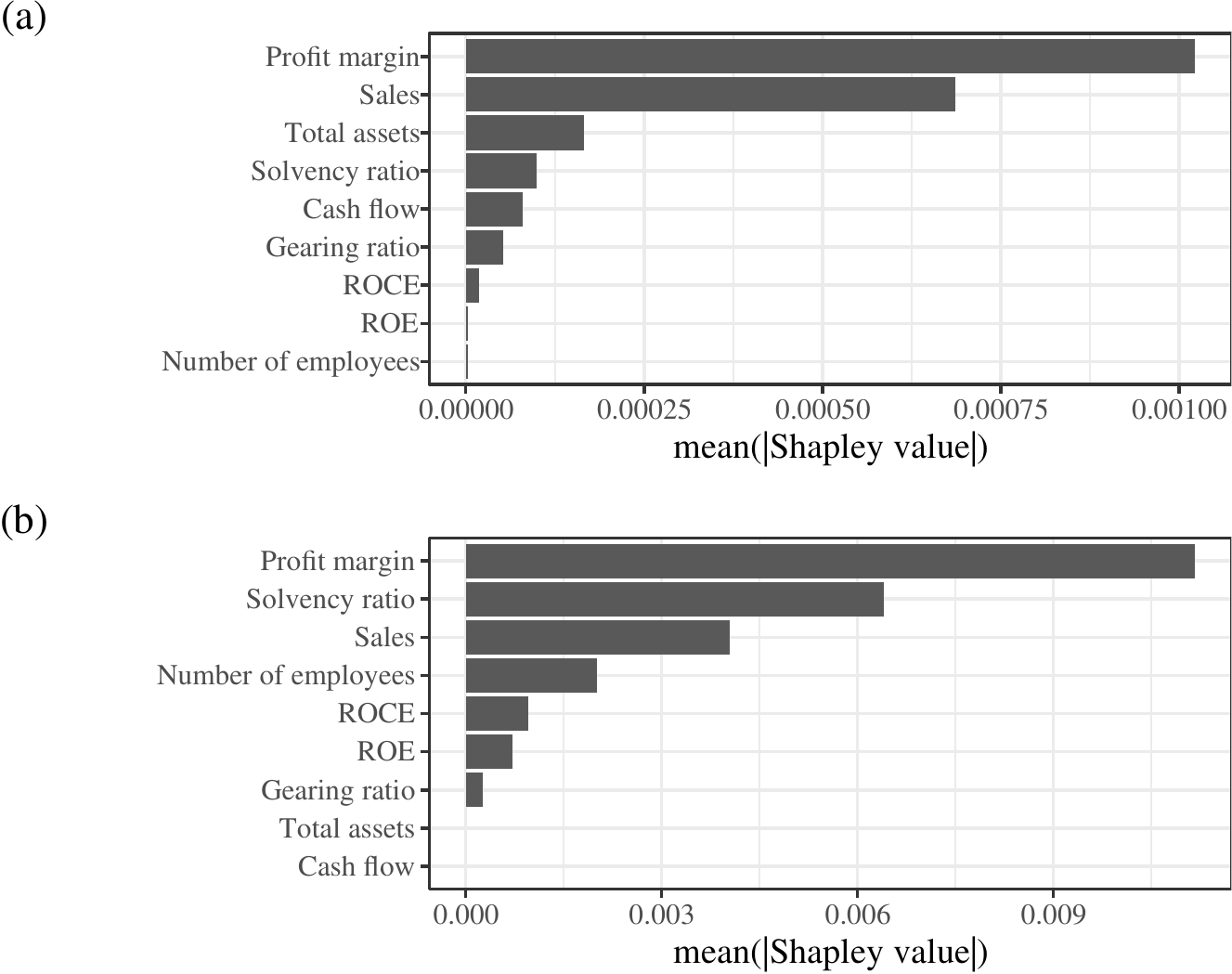}
	\caption{\label{fig:shapley}Global Shapley values for the Feedforward 
		Artificial Neural Network model (a) and the XGBoost model (b)}
\end{figure}

\begin{figure}
	\centering
	\subfloat[a][]{
		\includegraphics[width=0.80\textwidth]{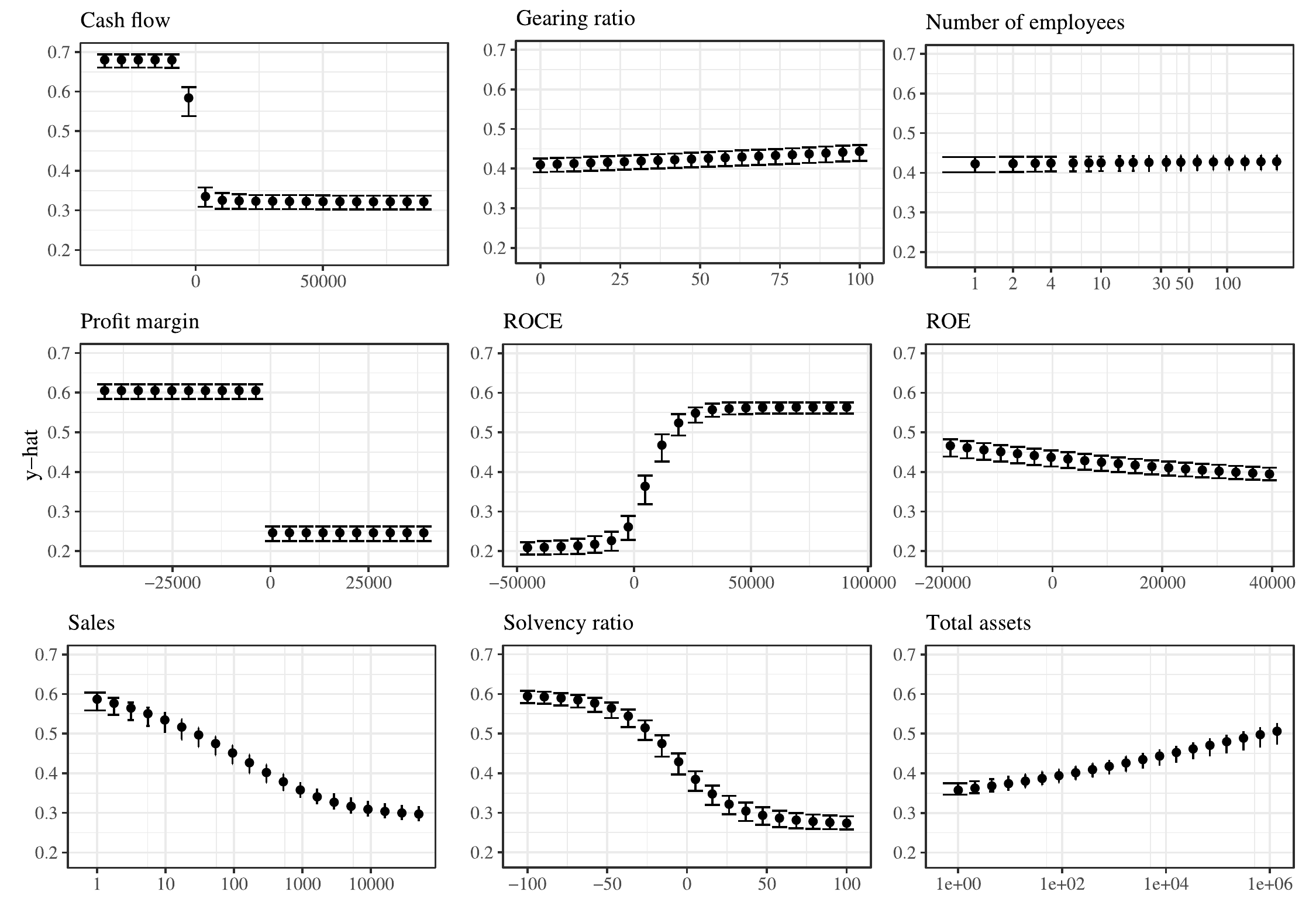}
	} \\
	\subfloat[b][]{
		\includegraphics[width=0.80\textwidth]{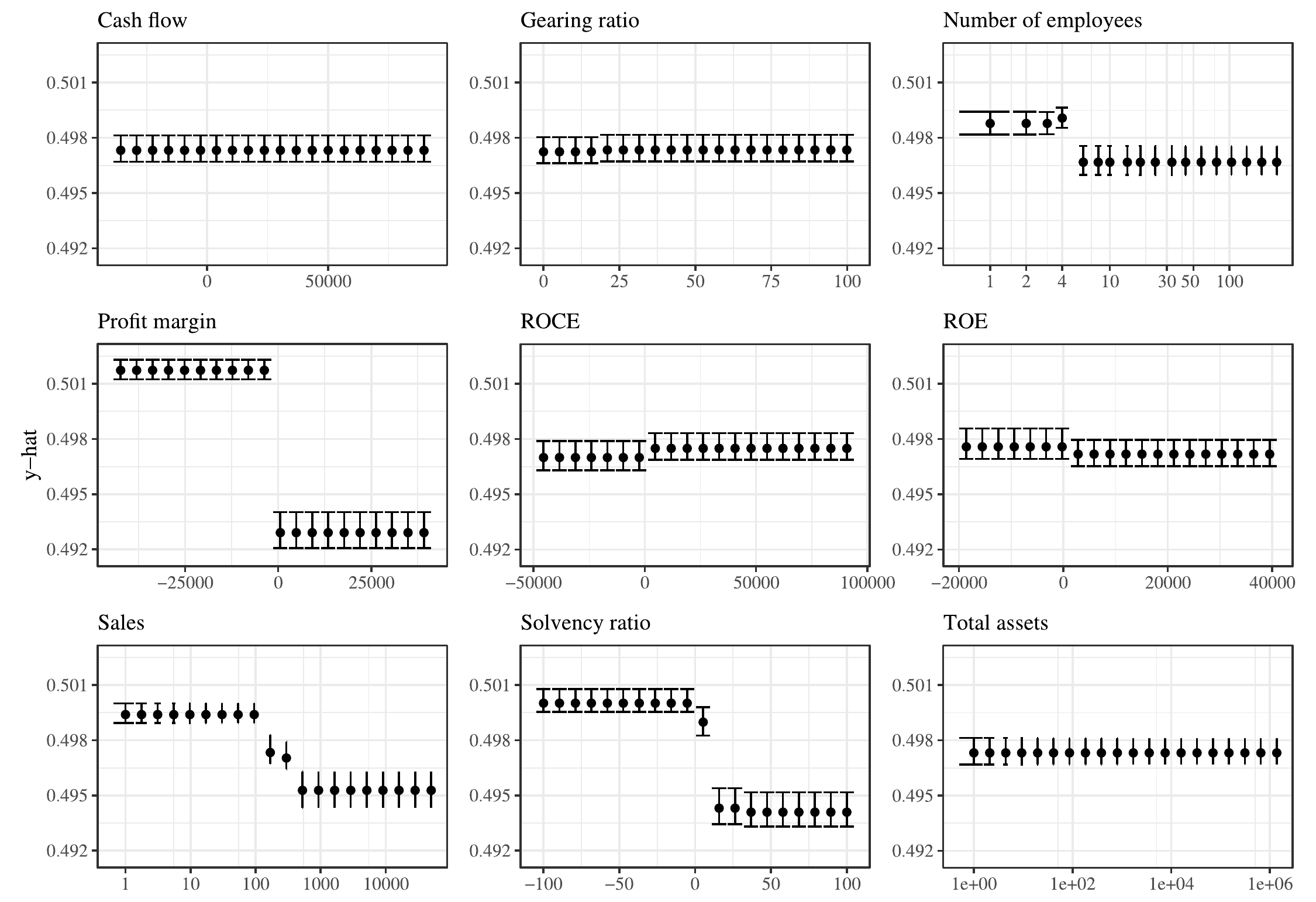}
	}
	\caption{
		Accumulated Local Effects of the FANN (a) and XGBoost (b) with related bootstrap 5\%-95\% confidence intervals. 
		The ALEs for Sales, Total Assets and Number of Employees are calculated on log-transformation of the variables but depicted on anti-log values to enhance interpretability.}\label{fig:alexgboost}
\end{figure}

The picture changes when it comes to black-box models. Global Shapley values 
indicate (figure \ref{fig:shapley}) that both FANN and XGBoost predictions are 
influenced mainly by Profit margin. 
This outcome is further 
clarified by the average change in the model output corresponding to increasing 
values of the variable, represented by ALEs (figure \ref{fig:alexgboost}).
The ALEs of either model show a downward sharp jump in the probability of 
default when 
moving from negative to positive values of Profit margin, with no further 
decrease in the probability of default as 
the ratio increases, revealing a clearly 
decreasing effect of this ratio on the probability of default, as previously 
found by \cite{altman2010sabato}, \cite{andreeva2016comparative} and \cite{petropoulos2019}.

The negative impact of Sales, already emerged in the white-box models, is confirmed to a minor extent by both FANN  and XGBoost (second and third important variable respectively according to Shapley values). 
However, the pattern of the estimated default probabilities for Sales is 
unlike: a smooth path with no evident plateauing effect in FANN and a first sudden decrease around 100.000 euros and a second drop around 316.000 euros in XGBoost.

A remarkable difference with respect to the white-box models are the 
sways of Total assets and the Number of employees. Total assets is the third 
important variable for FANN according to the Shapley values and seems to 
increase the probability of default judging from ALEs. On the contrary, the 
variable does not appear important in the prediction by XGBoost (Shapley value 
close to 0 and flat ALE). 
A positive impact of Total assets on the probability of default is anomalous, 
though shared by other scholars \citep{andreeva2016comparative}, in the light 
of our descriptive statistics and 
referring to the literature on firm demography, where exit is usually associated to less 
tangible assets \citep{michala2013}. 
This effect could be associated to the same found by other authors in the credit scoring literature. 
In that case a non-linear behaviour could be explained from the fact that creditors do pursue firms with larger assets with the hope to get back the money they have lent, whereas firms with low tangible assets are less worth being pursued \citep{altman2010sabato, el2016effect}.  

A somewhat opposite situation regards the Number of employees: FANN attributes 
scarce weight to this variable whereas XGBoost highlights its moderate impact 
(fourth important variable in the Shapley values) and a decrease in the 
probability of default around 5 employees. 
The XGBoost algorithm seems therefore able to capture separate and concordant
effects of two variables of firm size, one on the input and the other on the 
output side, in decreasing the probability of default, contrary to other 
empirical applications \citep{andreeva2016comparative}.

The Solvency ratio behaves similarly to Sales, for which the XGBoost shows a plateauing effect after 0 that the FANN does not point out. 
However, its importance, measured by the Shapley values, differs between the two algorithms since it is the second most relevant variable for XGBoost and the fourth relevant variable in the FANN.

The Cash flow, the third variable impacting on default according to white-box models, keeps a negative sign also in FANN, while it is not relevant into the XGBoost model (as in \cite{michala2013}). 
The Gearing ratio, ROCE and ROE are of a little consequence for XGBoost output 
and even less for the FANN according to the Shapley values and to overlapping 
bootstrap confidence intervals in Figure \ref{fig:alexgboost}, but for the 
FANN's ALEs plot that displays ROCE (however small its importance) as enhancing 
the probability of default, which is in line with part of the literature 
(\cite{calabrese2016bankruptcy} pointed out ROCE's positive effect). 
Another part of the literature instead 
found it non-significant \citep{ahelegbey2019, giudici2020}.

To summarize, blurry effects of one or more variables are encountered for the 
FANN model (Total assets and ROCE) and for all the white-box models (ROCE for 
all of them, Profit margin only for the Probit). 
Considering the prominent roles assigned by FANN to both Sales and Total assets, it seems 
that these two variables compensate one another in the wrong way, resulting in 
a the lowest correct classification of defaulted firms among the competing 
models. 

An interesting puzzle remains regarding the completely 
different ranking in 
the importance of variables 
according 
to white versus black-box models. Keeping performance in mind, we should 
consider 
what emerges from the interpretation of the XGBoost output, attributing the 
highest sensitivity achieved to an evaluation of the interplay among the 
variables which ends up more effective in predicting default.


\hypertarget{conclusions}{%
\section{Conclusions}\label{conclusions}}

  
Making an AI system interpretable allows external observers to understand 
its working and meaning, with the non-negligible consequence of making it 
usable in the practice: when a firm (or a customer) applies for a credit line, 
it has the right to know the eventual reasons for a refusal. AI driven 
decisions must be explained - as much as possible- to and understood by those 
directly and 
indirectly 
affected, in order to allow for contesting of such decisions. This issue has 
become extremely 
relevant since both  academicians and practitioners has progressively embraced 
ML modelling of firm default due to excellent performances 
\citep{ciampi2021rethinking} and, concurrently, Institutions have started to 
question the trustworthiness of - and set boundaries for - a safe use of AI in 
the interest of all involved \citep{EC2019}.  On the same 
time, using AI methods might grant larger amounts of credit and result in lower 
default rates \citep{moscatelli2020corporate}.  

Here we contribute to the literature on SMEs default by showing that the good 
performances in classification tasks obtained through ML models can and should 
be accompanied by a clear interpretation of the role and type of effect played 
by the variables involved. Our approach belongs to the post-hoc model-agnostic 
 interpretability methods that, differently from the ante-hoc 
techniques, enable the comparison among white and black-boxes on a common 
ground. 

Using a collection of relevant accounting indicators, widely employed in the 
literature, for all the Italian SMEs available in the dataset Orbis by BvD for 
2016, we have supplied an accurate prediction of default in 2017. Thanks to our 
research design, caring for imbalance among classes and cross-validation to 
select the most performing rules, we have achieved fair rates of correct 
classifications for all the models involved. However, focusing in particular on 
the correct rate of default classification, the XGBoost algorithm prevails 
over three white-box models and over the alternative ML model FANN.   

Interpretability was provided by means of Shapley values and ALEs, two recent 
model-agnostic techniques which measure the relative importance of the 
predictors and shape the predictor-outcome relationship respectively. The 
analysis of the XGBoost ALEs reveals that such complex models 
capture highly non-linear patterns as the effects of sales on the probability 
of default, account for separate effects of correlated measures and suggest 
also non-trivial risky thresholds: something that was not completely grasped by 
any standard discriminant models.  

We think that the examination of ALEs for models which are 
already ante-hoc interpretable in the traditional scheme of statistical 
significance 
is quite revealing, both methodologically and empirically speaking. The latter 
models' ALEs permits to add different shades to the variables' effects with 
respect to the standard parameter-pvalues' paradigm.   
Finally, the assessment of ALEs' variability is fundamental to check the  
 output robustness and to evaluate the soundness of results. 

With this paper we have showed that, assumed interpretability to be crucial for 
building 
and maintaining users’ trust in AI systems, their -potential- superiority in 
classification tasks needs not be anymore an alibi to hide the underlying 
mechanisms in black-boxes.

The relevancy of this approach could become definitely more important for  
default prediction based on alternative sources of data, such as web-scraped 
information, whose dimensionality
 and complexity require the power of ML models and whose interpretability is 
 even more puzzling. This, as well as applications to a more extensive basket 
 of traditional predictors, might represent a good ground for further 
 research.   

\hypertarget{acknowledgements}{%
\section{Acknowledgements}\label{acknowledgements}}

This project was possible thanks to the MIUR's project ``Department of Excellence'' and the Department of Department of Economics, Management and Statistics Big Data Lab which provided the virtual machine on which the analysis was carried out.

\bibliographystyle{model2-names}
\bibliography{creditrisk_v1.bib}

%
%

\end{document}